%% file: main.tex
\documentclass[a4paper,conference]{IEEEtran}

\ifCLASSINFOpdf
\else
\fi

\usepackage{amsmath}
\usepackage{times}
\usepackage{epsfig}
\usepackage{amssymb}
\usepackage{subcaption}
\usepackage{makecell}
\usepackage{enumitem}
\usepackage{tabularx}
\usepackage{url}
\usepackage{multirow}
\usepackage{pgfplots}
\usepackage{accents}
\usepackage{scalerel}
\usepackage{booktabs}
\usepackage{cleveref}
\usepackage{graphicx}

\input{Notation.tex}

\begin{document}

% title.
\title{Class Conditional Alignment for Partial Domain Adaptation}

\author{\IEEEauthorblockN{Mohsen Kheirandishfard,\, Fariba Zohrizadeh,\, and Farhad Kamangar}
\IEEEauthorblockA{
\\
Department of Computer Science and Engineering,\\
University of Texas at Arlington, USA\\
\{mohsen.kheirandishfard, fariba.zohrizadeh\}@mavs.uta.edu \;\;\;
farhad.kamangar@uta.edu}}

\maketitle
%%%%%%%%%%%%%%%%%%%%%%%%%%%%%%%%%%%%%%%%%%
%%%%%%%%%%%%%%%%%%%%%%%%%%%%%%%%%%%%%%%%%%
\begin{abstract}
Adversarial adaptation models have demonstrated significant progress towards transferring knowledge from a labeled source dataset to an unlabeled target dataset. Partial domain adaptation (PDA) investigates the scenarios in which the source domain is large and diverse, and the target label space is a subset of the source label space. The main purpose of PDA is to identify the shared classes between the domains and promote learning transferable knowledge from these classes. In this paper, we propose a multi-class adversarial architecture for PDA. The proposed approach jointly aligns the marginal and class-conditional distributions in the shared label space by minimaxing a novel multi-class adversarial loss function. Furthermore, we incorporate effective regularization terms to encourage selecting the most relevant subset of source domain classes. In the absence of target labels, the proposed approach is able to effectively learn domain-invariant feature representations, which in turn can enhance the classification performance in the target domain. Comprehensive experiments on three benchmark datasets Office-$31$, Office-Home, and Caltech-Office corroborate the effectiveness of the proposed approach in addressing different partial transfer learning tasks.
\end{abstract}
\IEEEpeerreviewmaketitle

%%%%%%%%%%%%%%%%%%%%%%%%%%%%%%%%%%%%%%%%%%
%%%%%%%%%%%%%%%%%%%%%%%%%%%%%%%%%%%%%%%%%%
\section{Introduction}
With the impressive power of learning representations, deep neural networks have shown superior performance in a wide variety of machine learning tasks such as classification \cite{szegedy2013deep,ren2015faster}, object detection \cite{girshick2015fast,ren2015faster,redmon2016you}, etc. These notable achievements heavily depend on the availability of large amounts of labeled training data. However, in many applications, collecting sufficient labeled data is either difficult or time-consuming. One potential solution to reduce the labeling consumption is to build an effective predictive model using readily-available labeled data from a different but related source domain. Such learning paradigm generally suffers from the distribution shift between the source and target domains, which in turn poses a significant difficulty in adapting the predictive model to the target domain tasks. In the absence of target labels, unsupervised domain adaptation (UDA) seeks to enhance the generalization capability of the predictive model by learning feature representations that are discriminative and domain-invariant \cite{pan2011domain,gong2012geodesic,baktashmotlagh2013unsupervised}. Various approaches have been proposed in the literature to tackle UDA problems by embedding domain adaptation modules in deep architectures \cite{yosinski2014transferable,donahue2014decaf,long2015learning,tzeng2015simultaneous,deng2018image,kazemi2018unsupervised} (see \cite{wang2018deep} for a comprehensive survey on deep domain adaptation methods). A line of research is developed to align the marginal distributions of the source and target domains through minimizing discrepancy measures such as maximum mean discrepancy \cite{tzeng2014deep,long2015learning}, central moment discrepancy \cite{zellinger2017central}, correlation distance \cite{sun2016return,sun2016deep}, etc. In this way, they can map both domains into the same latent space, which results in learning domain-invariant feature representations. Another strand of research is focused on designing specific distribution normalization layers which facilitate learning separate statistics for the source and target domains \cite{li2016revisiting,cariucci2017autodial}. More recently, some research studies have been carried out based on the generative adversarial networks \cite{goodfellow2014generative} that aim to alleviate the marginal disparities across the domains by adversarially learning domain-invariant feature representations which are indistinguishable for a discriminative domain classifier \cite{long2018conditional,pei2018multi,wang2019transferable}. 

\input{Page1-Figure}

Despite the efficacy of the existing UDA methods, their superior performance is mostly limited to the scenarios in which the source and target domains share the same set of labels. With the goal of considering more realistic and practical cases, \cite{cao2018partial} introduced partial domain adaptation (PDA) as a new adaptation scenario in which the target label space is a subset of the source label space. The main challenge in PDA is to identify and reject the source domain classes that do not appear in the target domain, known as \textit{outlier classes}, mainly because they may exert negative impacts on the overall transfer performance \cite{pan2010survey,cao2018san}. Addressing this challenge enables the PDA methods to transfer models trained on large and diverse labeled datasets (e.g. ImageNet) to small-scale unlabeled datasets from different but related domains.

In this paper, we propose a novel adversarial approach for partial domain adaptation which seeks to automatically reject the outlier source classes and improve the classification confidence on \textit{irrelevant samples}, i.e. the samples that are highly dissimilar across the domains. The existing PDA methods often align the marginal distributions between the domains in the shared label space. Different from these methods, we propose a novel adversarial architecture that matches class-conditional feature distributions by minimaxing a multi-class adversarial loss function. Moreover, we propose to boost the target domain classification performance by incorporating two novel regularization functions. The first regularizer is a row-sparsity term on the output of the classifier to promote the selection of a small subset of classes that are in common between the source and target domains. The second one is a minimum entropy term which increases the classifier confidence level in predicting the labels of irrelevant samples from both domains. We empirically observe that our proposed approach considerably improves the state-of-the-art performance for various partial domain adaptation tasks on three commonly-used benchmark datasets Office-$31$, Office-Home, and Caltech-Office.

%%%%%%%%%%%%%%%%%%%%%%%%%%%%%%%%%%%%%%%%%%
%%%%%%%%%%%%%%%%%%%%%%%%%%%%%%%%%%%%%%%%%%
\section{Related Work}
To date, various unsupervised domain adaptation (UDA) methods have been developed to learn domain-invariant feature representations in the absence of target labels. Some studies have proposed to minimize the maximum mean discrepancy between the features extracted from the source and target samples \cite{long2015learning,kazemi2018unsupervised,long2016unsupervised,yan2017mind,ghifary2014domain}. In \cite{sun2017correlation}, a correlation alignment (CORAL) method is developed that utilizes a linear transformation to match the second-order statistics between the domains. \cite{sun2016deep} presented an extension of the CORAL method that learns a non-linear transformation to align the correlations of layer activations in deep networks. Despite the practical success of the aforementioned methods in domain alignment, it is shown that they are unable to completely eliminate the domain shift \cite{donahue2014decaf,yosinski2014transferable}. Another line of work has proposed to reduce the discrepancy by learning separate normalization statistics for the source and target domains \cite{li2016revisiting,cariucci2017autodial}. \cite{li2016revisiting} adopts different batch normalization layers for each domain to align the marginal distributions. \cite{cariucci2017autodial} embeds domain alignment layers at different levels of a deep architecture to align the domain feature distributions to a canonical one. 

More recently, adversarial adaptation methods have been extensively investigated to boost the performance of UDA methods \cite{ganin2016domain,ghifary2016deep,bousmalis2017unsupervised,tzeng2017adversarial,long2018conditional}. The basic idea behind these methods is to train a discriminative domain classifier for predicting domain labels and a deep network for learning feature representations that are indistinguishable by the discriminator. By doing so, the marginal disparities between the source and target domains can be efficiently reduced, which results in significant improvement in the overall classification performance \cite{ganin2016domain,tzeng2017adversarial,pei2018multi}. Transferable attention for domain adaptation \cite{wang2019transferable} proposed an adversarial attention-based mechanism for UDA, which effectively highlights the transferable regions or images. \cite{zhang2018collaborative} introduced an incremental adversarial scheme which gradually reduces the gap between the domain distributions by iteratively selecting high confidence pseudo-labeled target samples to enlarge the training set. While the existing UDA models have shown tremendous progress towards reducing domain discrepancy, they mostly rely on the assumption of fully shared label space and generally align the marginal feature distributions between the source and target domains. This assumption is not necessarily valid in partial domain adaptation (PDA) which assumes the target label space is a subset of the source label space.

Great studies have been conducted towards the task of PDA to simultaneously promote positive transfer from the common classes between the domains and alleviate the negative transfer from the outlier classes \cite{cao2018san,cao2018partial,zhang2018importance}. Importance weighted adversarial nets \cite{zhang2018importance} develops a two-domain classifier strategy to estimate the relative importance of the source domain samples. Selective adversarial network (SAN) \cite{cao2018san} trains different domain discriminators for each source class separately to align the distributions of the source and target domains across the shared label space. Partial adversarial domain adaptation (PADA) \cite{cao2018partial} adopts a single adversarial network and incorporates class-level weights to both source classifier and domain discriminator for down-weighing the samples of outlier source classes. Example Transfer Network (ETN) \cite{cao2019learning} improves upon the PADA approach by introducing an auxiliary domain discriminator to quantify the transferability of each source sample.

%%%%%%%%%%%%%%%%%%%%%%%%%%%%%%%%%%%%%%%%%%%%%%%%%%%%%%%
Despite the efficacy of the existing PDA approaches in various tasks, they often align the marginal distributions of the shared classes between the domains without considering the conditional distributions \cite{cao2018partial,zhang2018collaborative,cao2019learning}. This may degenerate the performance of the model due to the negative transfer of irrelevant knowledge. To circumvent this issue, we utilize pseudo-labels for the target domain samples and develop a multi-class adversarial architecture to jointly align the marginal and class-conditional distributions (see \cref{tb:Page1} for more clarification). Inspired by \cite{xie2018learning}, we propose to align labeled source centroid and pseudo-labeled target centroid to mitigate the adverse effect of the noisy pseudo-labels. Similar to \cite{cao2018partial}, we incorporate class-level weights into our cost function to down-weight the contributions of the source samples belonging to the outlier classes. Furthermore, we introduce two novel regularization functions to promote the selection of a small subset of classes that are in common between the source and target domains and enhance the classifier confidence in predicting the labels of irrelevant samples from both domains.

%%%%%%%%%%%%%%%%%%%%%%%%%%%%%%%%%%%%%%%%%%%%%%%%%%%%%
\input{Model_fig.tex}
%%%%%%%%%%%%%%%%%%%%%%%%%%%%%%%%%%%%%%%%%%%%%%%%%%%%%
\section{Problem Formulation}
Let $\{(\xbf_{s}^{i}, \ybf_{s}^{i}){\}}_{i=1}^{n_{s}}$ be a set of $n_{\tav{s}}$ sample images collected $i.i.d$ from the source domain $\Dcal_{\tav{s}}$, where $\xbf_{\!s}^{i}$ denotes the $i^{\mathrm{th}}$ source image with label $\ybf_{s}^{i}$. Similarly, let $\{\xbf_{t}^{i}\}_{i=1}^{n_{t}}$ be a set of $n_{t}$ sample images drawn $i.i.d$ from the target domain $\Dcal_{t}$, where $\xbf_{t}^{i}$ indicates the $i^{\mathrm{th}}$ target image. To clarify the notation, let $\Xcal\!=\!\Xcal_{\tav{s}}\!\cup\!\Xcal_{\tav{t}}$ be the set of entire images captured from both domains, where $\Xcal_{\tav{s}}\!=\!\{\xbf_{\tav{s}}^{i}\}_{i=1}^{n_s}$ and $\Xcal_{\tav{t}}\!=\!\{\xbf_{\tav{t}}^{i}\}_{i=1}^{n_t}$. The UDA methods assume the source and target domains possess the same set of labels, denoted as $\Ccal_s$ and $\Ccal_t$, respectively. In the absence of target labels, the primary goal of the UDA methods is to learn transferable features 
that can reduce the shift between the marginal distributions of both domains. One promising direction towards this goal is to train a domain adversarial network \cite{tzeng2015simultaneous,ganin2016domain} consisting of a discriminator $G_{{d}}$ for predicting the domain labels, a feature extractor $G_{\!{f}}$ to learn domain-invariant feature representations for deceiving the discriminator, and a classifier $G_{{y}}$ that classifies the source domain samples. Training such adversarial network is equivalent to solving the following optimization problem
\begin{equation}
\begin{aligned}
\max_{\thetabf_{\!\tav{d}}}\,\min_{\thetabf_{\!\tav{y}},\!\;\!\thetabf_{\!\tav{f}}} ~~
& \frac{1}{n_{\tav{s}}}\!\sum_{
\begin{subarray}{c} \xbf^{i}\in\Xcal_{s} 
\end{subarray}} \!\! L_{\tav{y}}(G_{\tav{y}}(G_{\!\tav{f}}(\xbf^{\tav{i}};\thetabf_{\tav{f}}); \thetabf_{\tav{y}}), \ybf^{i}_{s})
\\
- & \frac{\lambda}{\,n\,} \sum_{\xbf^{i}\in\Xcal}  \! L_{\tav{d}}(G_{\tav{d}}(G_{\!\tav{f}}(\xbf^{\tav{i}};\thetabf_{\tav{f}}); \thetabf_{\tav{d}}), d^{\,i}),
\end{aligned}
\end{equation}
where $n \!=\! n_{\tav{s}}\!+n_{\tav{t}}$ denotes the total number of images, $\lambda\!>\!0$ is a regularization parameter, $\ybf^{i}_{s}$ is a one-hot vector denoting the class label of image $\xbf^{i}$, and $d^{\,i}\!\in\!\{0,1\}$ indicates its domain label. $L_{y}$ and $L_{d}$ are cross-entropy loss functions corresponding to the classifier $G_{y}$ and the domain discriminator $G_{d}$, respectively. Moreover, variables $\thetabf_{\!f}$, $\thetabf_{y}$, and $\thetabf_{\!d}$ are the network parameters associated with $G_{\!f}$, $G_{\!y}$, and $G_{\!d}$, respectively. For the brevity of notation, we drop the reference to the network parameters in the subsequent formulations.

%%%%%%%%%%%%%%%%%%%%%%%%%%%%% Partial domain
As noted earlier, standard domain adaptation approaches assume that the source and target domains possess the same label space, i.e. $\Ccal_{s}=\Ccal_{t}$. This assumption may not be fulfilled in a wide range of practical applications in which $\Ccal_{s}$ is large and diverse (e.g., ImageNet) and $\Ccal_{t}$ only contains a small subset of the source classes (e.g., Office-31), i.e. $\Ccal_{\tav{t}}\!\subset\! \Ccal_{\tav{s}}$. In such scenarios, it is hard to identify the shared label space between the domains since target labels and target label space $\Ccal_{t}$ are 
unknown during the training procedure. Under this condition, matching the marginal distributions may not necessarily facilitate the classification task in the target domain and a classifier with adaptation may perform worse than a standard classifier trained on the source samples. This is attributed to the adverse effect of transferring information from the outlier classes $\Ccal_{\tav{s}}\!\setminus\!\Ccal_{\tav{t}}$ \cite{cao2018san,cao2018partial}. Hence, the primary goal in partial domain adaptation is to identify and reject the outlier classes and simultaneously align the conditional distributions of the source and target domains across the shared label space. One of the well-established works toward this goal is Partial Adversarial Domain Adaptation (PADA) \cite{cao2018partial} which highlights the shared classes and reduces the importance of the outlier classes via the following weighting procedure
%%%%%%%%%%%%%%%%%%%%%%%%%%%%%%%%%%%%%%%%%%%%%%%%%%%%%%%%%%%%%%%%
\begin{equation}\label{eq:weights_pada}
\begin{aligned}
\gammabf = \frac{1}{n_{t}}\sum_{i=1}^{n_{t}}\hat{\ybf}^{i}_{t},
\end{aligned}
\end{equation}
%%%%%%%%%%%%%%%%%%%%%%%%%%%%%%%%%%%%%%%%%%%%%%%%%%%%%%%%%%%%%%%%
where $\hat{\ybf}^{i}_{t}=G_{y}(G_{\!f}(\xbf^{i}_{t}))$ denotes the output of the classifier $G_{y}$ to the target sample $\xbf^{i}_{t}$ and it can be considered as a probability distribution over the source label space $\Ccal_{s}$. The weight vector $\gammabf$ is further normalized as $\gammabf\leftarrow\gammabf\!\setminus\!\max({\gammabf})$ to demonstrate the relative importance of the classes. The weights associated with the outlier classes are expected to be much smaller than that of the shared classes, mainly because the target samples are significantly dissimilar to the samples belonging to the outlier classes. Ideally, $\gammabf$ is expected to be a vector whose elements are non-zero except those corresponding to the outlier classes. Given that, PADA proposed to down-weigh the contributions of the source samples belonging to the outlier classes $\Ccal_{s}\!\setminus\!\Ccal_{t}$ by adding the class-level weight vector $\gammabf$ to both source classifier $G_{y}$ and domain discriminator $G_{d}$. Therefore, the objective of PADA can be formulated as follows
\begin{equation}\label{eq:Total}
\begin{aligned}
\underset{\begin{subarray}{l} \thetabf_{\!\tav{d}}\end{subarray}}{\text{max}}
\,\underset{\begin{subarray}{l} \thetabf_{\!\tav{y}},\!\;\!\thetabf_{\!\tav{f}}\end{subarray}}{\text{min}} ~~
& \frac{1}{n_{\tav{s}}} \sum_{
\begin{subarray}{c} \xbf^{i}\in\Xcal_{s} 
\end{subarray}} \!\! \gamma_{c_{i}}\, L_{\tav{y}}(\msh G_{\tav{y}}(\msh G_{\!\tav{f}}(\xbf^{\tav{i}})\msh), \ybf^{i})
\\
- & \frac{\lambda}{n_{s}} \sum_{\xbf^{i}\in\Xcal_{s}}  \!\! \gamma_{c_{i}}\, L_{\tav{d}}(\msh G_{\tav{d}}(\msh G_{\!\tav{f}}(\xbf^{\tav{i}})\msh), d^{\,i})
\\
- & \frac{\lambda}{n_{t}} \sum_{\xbf^{i}\in\Xcal_{t}}  \! L_{\tav{d}}(\msh G_{\tav{d}}(\msh G_{\!\tav{f}}(\xbf^{\tav{i}})\msh), d^{\,i}),
\end{aligned}
\end{equation}
where scalar $\gamma_{c_{i}}$ denotes the class weight of sample $\xbf^{\tav{i}}$ and $c_{i}\!=\!\mathrm{argmax}_{j}\,y^{i}_{j}$ indicates the index of the largest element in vector $\ybf^{i}$. 

%%%%%%%%%%%%%%%%%%%%%%%%%%%%%%%%%%%%%%%%%%
%%%%%%%%%%%%%%%%%%%%%%%%%%%%%%%%%%%%%%%%%%
\section{Proposed Method}
Although the weighting scheme \cref{eq:weights_pada} is able to effectively match the marginal distributions of the source and target domains in the shared label space, there is no guarantee that the corresponding class-conditional distributions can also be drawn close. This may significantly degenerate the performance of the model due to the negative transfer of irrelevant knowledge. To circumvent this issue, we introduce a novel adversarial architecture to jointly align the marginal and class-conditional distributions in the shared label space. The proposed model adopts a multi-class discriminator $\tilde{G_{d}}$, parameterized by $\tilde{\thetabf}_{d}$, to classify the feature representations $G_{\!f}(\xbf^{i})$ into $2\times|\Ccal_{s}|$ categories, where the first and the last $|\Ccal_{s}|$ categories respectively correspond to the probability distribution over the source label space $\Ccal_{s}$ and target label space $\Ccal_{t}$ ($\Ccal_{t}\!\subset\!\Ccal_{s}$). We propose to train the discriminator $\tilde{G_{d}}$ with the following objective function
%====================
\begin{equation}\nonumber
\begin{aligned}
\tilde{\Lcal}_{d}(\thetabf_{\!f}, \tilde{\thetabf}_{d}) \!=\! - &\frac{1}{n_{s}}\! \sum_{\xbf^{i}\in\Xcal_{s}} \! \gamma_{c_{i}} L_{{d}}(\msh \tilde{G}_{{d}}(\msh G_{\!{f}}(\msh\xbf^{i}\msh)\msh), \tilde{\dbf}^{\,i}\msh)
\\
- &\frac{1}{n_{t}}\! \sum_{\xbf^{i}\in\Xcal_{t}} \! L_{d}(\msh \tilde{G}_{d}(\msh G_{\!f}(\msh\xbf^{i}\msh)\msh), \tilde{\dbf}^{\,i}\msh),
\end{aligned}
\end{equation}
%====================
%====================
where vector $\tilde{\dbf}^{\,\,\!i}\in\Rbb^{2\times|\Ccal_{s}|}$ is defined as the domain-class label of sample point $\xbf^{i}$. Due to the lack of class labels for the target samples, we set $\tilde{\dbf}^{\,\,\!i}$ to $[\ybf^{i}_{s},\mathbf{0}]$ if $\xbf^{i}\!\in\!\Xcal_{s}$ and use  $[\mathbf{0}, \tilde{\ybf}_{t}^{i}]$ if $\xbf^{i}\!\in\!\Xcal_{t}$, where $\tilde{\ybf}_{t}^{i}$ corresponds to the pseudo-label generated by classifier $G_{y}$ and is given by $\Tilde{\ybf}_{{t}}^{i} \!=\! \mathrm{argmax}_{c}\, \ebf_{c}^{\!\top} G_{y}( G_{\!f}(\xbf_{t}^{i})),
$ where $\{\ebf_{{c}}\}_{\tav{c=1}}^{{|\Ccal_{\tav{s}}|}}$ denotes the standard unit basis in $\Rbb^{|{\Ccal_{\tav{s}}}|}$. Moreover, the negative transfer can be efficiently alleviated by incorporating the weight vector $\gammabf$ into the loss $\tilde{\Lcal}_d$ which results in selecting out the source samples belonging to the outlier label space $\Ccal_{\tav{s}}\!\setminus\!\Ccal_{\tav{t}}$. It is noteworthy to mention that the direct use of pseudo-labels may degrade the classification performance as the pseudo-labels are predicted by the classifier and hence they may be noisy and inaccurate. Many literature methods leverage the theory of domain adaptation \cite{ben2007analysis} to present error analysis and derive certain bounds on the error introduced by incorporating the pseudo-labels \cite{saito2017asymmetric,xie2018learning}. These analysis are not generally applicable to the problem of partial domain adaptation as they mainly rely on the assumption that the source and target domains possess the same set of labels.

With the proposed multi-class adversarial loss $\tilde{\Lcal}_{d}$, the key challenge is how to tackle the uncertainty in pseudo-labels. One promising approach to mitigate the adverse effect of falsely-pseudo-labeled target samples is to align labeled source centroids and pseudo-labeled target centroids in the feature space \cite{xie2018learning}. However, this approach hardly fits the partial domain adaptation scenario in which the target label space is a subset of source label space. We propose to modify the aforementioned approach by incorporating weight vector $\gammabf$ to highlight the mismatch between the centroids of the shared classes. Hence, the weighted centroid alignment loss function can be formulated as
%====================
\begin{equation}\nonumber
\begin{aligned}
\Lcal_{\tav{c}}( \thetabf_{\!f},\, \thetabf_{y} ) = &\sum_{i=1}^{|\Ccal_{s}|} {\gamma_{i}}\,{\lVert{\Mbf_{\!s}^{i}-\Mbf_{\!t}^{i}}\rVert}^{2}_{2},
\end{aligned}
\end{equation}
%====================
where $\Mbf_{\!s}^i$ and $\Mbf_{\!t}^i$ respectively denote the feature centroids for the $i^{th}$ class in the source and target domains. These vectors are computed via the following formulas
%====================
\begin{equation}\nonumber
\begin{aligned}
\Mbf_{\!s}^i=\frac{1}{\lvert\Ocal_i\rvert}\sum_{\xbf^i\in\Ocal_i}\!G_{\!f}(\xbf^{i}),\quad \Mbf_{\!t}^i=\frac{1}{\lvert\tilde{\Ocal}_i\rvert}\sum_{\xbf^i\in\tilde{\Ocal}_i}\!G_{\!f}(\xbf^{i}),
\end{aligned}
\end{equation}
%====================
where $\Ocal_i$ is the set of source samples belonging to the $i^{th}$ class and $\tilde{\Ocal}_i$ denotes the set of target samples assigned to the $i^{th}$ class.

In what follows, we propose two novel regularization functions to derive more discriminative class weights and to increase the confidence level of the classifier in predicting the labels of the irrelevant samples across both domains.
%%%%%%%%%%%%%%%%%%%%%%%%%%%%%%%%%%%%%%%%%%%%%%%%%%%%%%%%%%%%%%%%

Motivated by the assumption that the target samples are dissimilar to the samples of the outlier classes, we propose a row-sparsity regularization term that promotes the selection of a small subset of source domain classes that appear in the target domain. This, in turn, encourages the weight vector $\gammabf$ to be a vector of all zeros except for the elements corresponding to the shared classes. This selection regularization can be formulated as follows 
%%%%%%%%%%
\begin{equation}\nonumber
\begin{aligned}
\Lcal_{\sm{\infty}{6}}(\thetabf_{\!f},\thetabf_{y})\msh=\msh \frac{1}{|\Ccal_{s}|} 
\big\lVert {G_{\msh\tav{y}}(\msh G_{\!\tav{f}}(\xbf^{\tav{1}}_{t})\msh),\dots,\msh G_{\msh\tav{y}}(\msh G_{\!\tav{f}}(\xbf^{{\lvert \mathcal{X}_{t}\rvert}}_{t})\msh) }{\big{\rVert}}_{1,\sm{\infty}{6}},
\end{aligned}
\end{equation}
%%%%%%%%%
where $\lvert .\rvert$ denotes the cardinality of its input set and ${\lVert . \rVert}_{1,\sm{\infty}{6}}$ computes the sum of the infinity norms of the rows of an input matrix. To illustrate, for an arbitrary matrix $\Abf=[\abf_1|\abf_2|\dots|\abf_n]^{\!\top}\in\Rbb^{n\times m}$, scalar $\lVert\abf_i{\rVert}_{\infty}$ denotes the maximum absolute value of $i^{th}$ row. Therefore, regularization term ${\lVert \Abf \rVert}_{1,\sm{\infty}{6}}=\sum_{i=1}^n\lVert\abf_i{\rVert}_{\infty}$ promotes sparsity on the maximum absolute value of each row which in turn leads to some zero rows in matrix $\Abf$. 

The regularization term $\Lcal_{\sm{\infty}{6}}$ takes into consideration the relation between the entire target samples and encourages the classifier to generate a sparse output vector with its non-zero entries located at certain indices correspond to the classes shared between the domains. Notice that this regularization term does not directly enforce a specific number of classes to be chosen but rather promotes the network to select a subset of source domain classes.

%%%%%%%%%%%%%%%%%%%%%%%%%%%%%%%%%%%%%%%%%%%%%%%%%%%%%%%%%%%%%%%%%%%
Besides the outlier classes, the irrelevant samples are inherently less transferable and they may significantly degrade the target classification performance in different PDA tasks. To reduce the negative effect of irrelevant samples in the training procedure, we propose to leverage the following entropy minimization term
\begin{equation}\nonumber
\begin{aligned}
\hspace{-0.2cm}\Lcal_{e}(\thetabf_{\!f},\thetabf_{y})\!= &\frac{1}{n_s} \!\sum_{\xbf^{i}\in\Xcal_{s}} \!\!\!\!
\gamma_{c_{i}} L_{y}^{e}(\msh G_{{y}}(\msh G_{\!{f}}(\xbf^{{i}})\msh)\msh)
\\
+&\frac{1}{n_t}
\sum_{\xbf^{i}\in\Xcal_{t}} \!\!
L_{y}^{e}(G_{{y}}(G_{\!{f}}(\xbf^{{i}})\msh),
\end{aligned}
\end{equation}
where $L_{y}^{e}$ is the entropy loss functions corresponding to the classifier $G_{y}$. Generally, regularization $\Lcal_e$ encourages the classifier to produce vectors with one dominant element denoting the label (or pseudo-label) of samples. This, in turn, enhances the performance of the feature extractor and helps to learn more transferable features for classification. Moreover, weight vector $\gamma$ is incorporated to highlight the importance of samples belonging to the shared classes.

%%%%%%%%%%%%%%%%%%%%%%%%%%%%%%%%%%%%%%%%%%%%%%%%%%%%%%
By combining the aforementioned loss functions, training our proposed model is equivalent to solving the following minimax saddle point optimization problem
%%%%%%%%%%%%%%%%%%%%%%%%%%%%%%%%%%%%%%%%%%%%%%%%%%%%%%
\begin{equation}
\begin{aligned}
\underset{\begin{subarray}{l} \tilde{\thetabf}_{d}\end{subarray}}{\text{max}}~
\,\underset{\begin{subarray}{l} \vphantom{\tilde{\thetabf}} \thetabf_{{y}}, \thetabf_{{f}}\end{subarray}}{\text{min}} ~~
&\; \frac{1}{n_{{s}}}\! \sum_{\begin{subarray}{c} \xbf^{i}\in\Xcal_{s} \end{subarray}} \!\! \gamma_{c_{i}}\, L_{\tav{y}}(\msh G_{{y}}(\msh G_{\!{f}}(\xbf^{{i}})\msh), \ybf^{i})
\\
 & + \lambda \, \tilde{\Lcal}_{d}\,(\thetabf_{\!f}, \tilde{\thetabf}_{d}) + \Lcal_{{c}}(\thetabf_{\!f}, \thetabf_{y})
\\[1.7ex]
 & + \mu \, \Lcal_{\sm{\infty}{6}}(\thetabf_{\!f},\thetabf_{y}) + \zeta  \, \Lcal_{e}(\thetabf_{\!f},\thetabf_{y}), 
\end{aligned}
\end{equation}
where $\lambda$, $\mu$, and $\zeta $ are positive hyperparameters to control the contribution of each loss component. 
%%%%%%%%%%%%%%%%%%%%%%%%%%%%%%%%%%%%%%%%%%
%%%%%%%%%%%%%%%%%%%%%%%%%%%%%%%%%%%%%%%%%%
\section{Experiments}
This section evaluates the efficacy of our approach, named {\PaperName}, through conducting empirical experiments on two widely used benchmark datasets for partial domain adaptation (PDA) problem. The experiments are performed on different PDA tasks in an unsupervised setting where neither the target labels nor the target label space is available. In what follows, we give more explanations about the datasets, the PDA tasks, and the network hyperparameters used in our experiments.
%%%%%%%%%%%%%%%%%%%%%%%%%%%%%%%%%%%%%%%%%%
\subsection{Setup}
\input{Fig_office_home.tex}
\input{Table_office.tex}

\input{Table_caltech_office.tex}
%%%%%%%%%%%%%%%%%%%%%%%%%%%%%%%%%%%%%%%%%%
\noindent \textbf{Dataset:} We evaluate the performance of {\PaperName} on two commonly used datasets for the task of partial domain adaptation: Office-$31$ \cite{saenko2010adapting}, Office-Home \cite{venkateswara2017deep}, and Caltech-Office. Office-$31$ object dataset consists of $4,652$ images from $31$ classes, where the images are collected from three different domains:  \textit{Amazon} (\textbf{A}), \textit{Webcam} (\textbf{W}), and \textit{DSLR} (\textbf{D}). We follow the procedure presented in the literature \cite{cao2018partial,cao2019learning} to transfer knowledge from a source domain with $31$ classes to a target domain with $10$ classes. The results are reported as the average classification accuracy of the target domain over five independent experiments across six different PDA tasks: \textbf{A} $\rightarrow$ \textbf{W}, \textbf{W} $\rightarrow$ \textbf{A}, \textbf{D} $\rightarrow$ \textbf{W}, \textbf{W} $\rightarrow$ \textbf{D}, \textbf{A} $\rightarrow$ \textbf{D}, and \textbf{D} $\rightarrow$ \textbf{A}. 

Office-Home is a more challenging dataset that contains $15,500$ images collected from four distinct domains: \textit{Art} (\textbf{Ar}), \textit{Clipart} (\textbf{Cl}), \textit{Product} (\textbf{Pr}), and \textit{Real-World} (\textbf{Rw}), where each domain has $65$ classes. Example images from this dataset are provided in \cref{fig:finalResults1}. Following the procedure presented in \cite{cao2018partial,cao2019learning}, we aim to transfer information from a source domain containing $65$ classes to a target domain with $25$ classes. The results on this dataset are also reported as the average classification accuracy of the target domain over five independent experiments across twelve pairs of source-target adaptation tasks: \textbf{Ar} $\rightarrow$ \textbf{Cl}, \textbf{Ar} $\rightarrow$ \textbf{Pr},
\textbf{Ar} $\rightarrow$ \textbf{Rw}, \textbf{Cl} $\rightarrow$ \textbf{Ar}, \textbf{Cl} $\rightarrow$ \textbf{Pr}, \textbf{Cl} $\rightarrow$ \textbf{Rw}, \textbf{Pr} $\rightarrow$ \textbf{Ar}, \textbf{Pr} $\rightarrow$ \textbf{Cl}, \textbf{Pr} $\rightarrow$ \textbf{Rw}, \textbf{Rw} $\rightarrow$ \textbf{Ar}, \textbf{Rw} $\rightarrow$ \textbf{Cl}, and \textbf{Rw} $\rightarrow$ \textbf{Pr}.

Caltech-Office \cite{gong2012geodesic} is constructed from Caltech-$256$ \cite{griffin2007caltech} dataset as the source domain and Office-$31$ as the target domain. Following the procedure in \cite{cao2018san}, we consider the ten categories shared by Caltech-$256$  and Office-$31$ as the shared label space. Denoting the source domain as \textbf{C}, the result on Caltech-Office dataset are reported as the average classification accuracy of the target domain over five independent experiments across three pairs of source-target adaptation tasks: \textbf{C}$\rightarrow$\textbf{W}, \textbf{C}$\rightarrow\,$\textbf{A}, and \textbf{C}$\rightarrow\,$\textbf{D}.

We follow the standard evaluation protocols for partial domain adaptation \cite{cao2018san,cao2018partial} and compare the performance of {\PaperName} against several deep transfer learning methods: 
Reverse Gradient (RevGrad) \cite{ganin2015unsupervised}, Domain Adversarial Neural Network ({DANN}) \cite{ganin2016domain}, Residual Transfer Networks ({RTN}) \cite{long2016unsupervised}, Adversarial Discriminative Domain Adaptation ({ADDA}) \cite{tzeng2017adversarial}, Importance Weighted Adversarial Nets (IWAN) \cite{zhang2018importance}, Multi-Adversarial Domain Adaptation (MADA) \cite{pei2018multi} , Selective Adversarial Network ({SAN}) \cite{cao2018san}, Partial Adversarial Domain Adaptation ({PADA}) \cite{cao2018partial}, and Example Transfer Network (ETN) \cite{cao2019learning}. Moreover, in order to demonstrate the efficacy brought by different components of the proposed PDA model, we conduct an ablation study by evaluating three variants of {\PaperName}: $\text{\PaperName}_{\infty}$ is a variant of {\PaperName} without incorporating the selection regularization term $\Lcal_{\infty}$, $\text{\PaperName}_{e}$ denotes a variant without considering $\Lcal_{e}$, and $\text{\PaperName}_{d,c}$ is a variant with a binary discriminator and without considering the weighted centroids alignment term $\Lcal_{c}$.

\noindent \textbf{Parameter:} We use PyTorch to implement {\PaperName} and adopt ResNet-$50$ \cite{he2016deep} model pre-trained on ImageNet \cite{russakovsky2015imagenet}, as the backbone for the network $G_{\!f}$. We fine-tune the entire feature layers and apply back-propagation to train the domain discriminator $\tilde{G}_d$ and the classifier $G_{y}$. Since parameters $\thetabf_{y}$ and $\tilde{\thetabf}_{d}$ are trained from scratch, their learning rates are set to be $10$ times greater than that of $\thetabf_{\!f}$. To solve the minimax problem \cref{eq:Total}, we use mini-batch stochastic gradient descent (SGD) with a momentum of $0.95$ and the learning rate is adjusted during SGD by: $\eta = \frac{\eta_{\tav{0}}}{(1+\alpha\times\rho)^{\beta}}$ where $\eta_{\tav{0}} = 10^{\smallMinus 2}$, $\alpha=10$, $\beta=0.75$, and $\rho$, denoting the training progress, linearly changes from $0$ to $1$ \cite{ganin2016domain,cao2018partial}. We use a batch size $b = 72$ with $36$ samples for each domain. Parameter $\mu$ is set to $0.1$ for Office-$31$, Office-Home, and Caltech-Office datasets. Notice that since the classifier is not appropriately trained in the first few epochs, the value of $\mu$ can be gradually increased from $0$ to $0.1$. Other hyper-parameters are tuned by importance weighted cross validation \cite{sugiyama2007covariate} on labeled source samples and unlabeled target samples.

As we use mini-batch SGD for optimizing our model, categorical information in each batch is usually inadequate for obtaining an accurate estimation of the source and target centroids. This in turn may adversely affect the alignment performance. To mitigate this issue, we align the moving average centroids of the source and target classes in the feature space (with coefficient $0.7$) rather than aligning the inaccurate centroids obtained in each iteration.
%%%%%%%%%%%%%%%%%%%%%%%%%%%%%%%%%%%%%%%%%%
\input{tsne_fig}
\input{Convergance_Table}
%%%%%%%%%%%%%%%%%%%%%%%%%%%%%%%%%%%%%%%%%%
\subsection{Results}
The target domain classification accuracy for various methods on six PDA tasks of Office-$31$ dataset, twelve PDA tasks of Office-Home dataset, and three PDA tasks of Caltech-Office dataset are reported in Tables \ref{tab:Office_31}, \ref{tab:office_home}, and \ref{tab:caltech_Office}. The entire results are reported based on the ResNet-$50$ and the scores of the competitor methods are directly collected from \cite{cao2018san} and \cite{cao2019learning}. 

Observe that unsupervised domain adaptation methods such as ADDA, DANN, and MADA have exhibited worse performance than the standard ResNet-$50$ on some PDA tasks in both datasets Office-$31$ and Office-Home. This can be attributed to the fact that these methods aim to align the marginal distributions across the domains and hence are prone to the negative transfer introduced by the outlier classes. On the other hand, the partial domain adaptation methods, such as PADA, SAN, IWAN, ETN, and {\PaperName}, achieve promising results on most of the PDA tasks since they leverage different mechanisms to highlight a subset of samples that are more transferable across both domains. 

Among the competing partial domain adaptation approaches in Tables \ref{tab:Office_31}, \ref{tab:office_home}, and \ref{tab:caltech_Office}, SAN is the only approach that seeks to directly aligns the conditional distributions of the source and target domains. However unlike {\PaperName}, SAN uses a different architecture with $|\Ccal_{s}|$ class-wise domain discriminators to identify the domain-class label of each sample. As reported in Tables \ref{tab:Office_31}, \ref{tab:office_home} and \ref{tab:caltech_Office}, {\PaperName} outperforms SAN with a large margin in all PDA tasks on both Office-$31$ and Office-Home datasets. Moreover, {\PaperName} requires fewer parameters compared to SAN. This in turn demonstrates the efficiency and efficacy of the proposed class-conditional model.

The results in Table \ref{tab:Office_31} indicate that {\PaperName} outperforms the competing methods on most of the PDA tasks from Office-$31$ dataset. In particular, {\PaperName} achieves considerable improvement on \textbf{A} $\rightarrow$ \textbf{W} and \textbf{A} $\rightarrow$ \textbf{D} tasks. It also increases the average accuracy of all tasks by almost $1.36\%$. Moreover, Table \ref{tab:office_home} shows that {\PaperName} outperforms other PDA approaches with a large margin on five pairs of source-target adaptation tasks: \textbf{Ar} $\rightarrow$ \textbf{Pr}, \textbf{Ar} $\rightarrow$ \textbf{Rw}, \textbf{Cl} $\rightarrow$ \textbf{Ar}, \textbf{Cl} $\rightarrow$ \textbf{Pr}, and \textbf{Rw} $\rightarrow$ \textbf{Ar}. The results reported in Table \ref{tab:caltech_Office} indicate that {\PaperName} outperform all comparison
methods on all the tasks even though the number of the outlier classes ($|\,\Ccal_{\tav{s}}\!\setminus \Ccal_{\tav{t}}\,|$) is much larger than that of the shared classes ($|\Ccal_{\tav{t}}|$). The numerical results provided in Tables \ref{tab:Office_31}, \ref{tab:office_home}, and \ref{tab:caltech_Office} corroborate {\PaperName} can effectively align the class-conditional distribution, mitigate transferring knowledge from the outlier source classes, and promote positive transfer between the domains in the shared label space. 

Furthermore, we perform an ablation study to evaluate the efficacy brought by different components of the proposed PDA model. We consider PADA as a baseline variant of {\PaperName} with binary domain discriminator $G_{d}$ and without regularization terms $\Lcal_{c}$, $\Lcal_{\tav{\infty}}$, and $\Lcal_{e}$. The results are reported in Table \ref{tab:ablation_study} and they reveal interesting observations. $\text{\PaperName}_{d,c}$ outperforms PADA in most of the tasks, which highlights the importance of the incorporated regularization terms $\Lcal_{\tav{\infty}}$ and $\Lcal_{e}$ in rejecting the outlier source samples. Moreover, we can see that both variants $\text{\PaperName}_{\tav{\infty}}$ and $\text{\PaperName}_{e}$ improved the accuracy of the original baseline, which corroborate the efficacy of our class-conditional domain discriminator $\tilde{G_{d}}$. Overall, observe that different components of the proposed method bring complimentary information into the model and they have contribution in achieving the state-of-the-art classification results.

\vspace{1mm}
\noindent\textbf{Visualization:} To better demonstrate the ability of the proposed method in aligning the feature distributions in the shared label space, we visualize the bottleneck representations learned by SAN, PADA, ETN, and {\PaperName} on task \textbf{A (31 classes)} $\rightarrow$ \textbf{W (10 classes)} using t-SNE embedding \cite{van2013barnes} (Shown in \cref{fig:tsne}). It is desired to embed the source and target sample points of the same class close together while keeping embeddings from different classes far apart. Observe that {\PaperName} is able to effectively discriminate the classes shared between the domains, while minimizing the distance between the same classes in both domains.
%%%%%%%%%%%%%%%%%%%%%%%%%%%%%%%%%%%%%%%
\input{Table_ablation.tex}
%%%%%%%%%%%%%%%%%%%%%%%%%%%%%%%%%%%%%%%

\vspace{1mm}
\noindent\textbf{Convergence Performance:} 
To highlight other advantages of our approach, we compare the test error rate obtained by {\PaperName} against various methods SAN, PADA, and ETN on partial domain adaptation task  \textbf{A (31 classes)} $\rightarrow$ \textbf{W (10 classes)}, from Office dataset. \cref{tb:convergance} illustrates the convergence behavior of the test errors in $15,000$ iterations. Each curve is obtained by averaging over $5$ independent runs for the entire test samples. Observe that comparing to the competitor methods, {\PaperName} not only converges very quickly but also achieves lower error rate. 

%%%%%%%%%%%%%%%%%%%%%%%%%%%%%%%%%%%%%%%%%%
%%%%%%%%%%%%%%%%%%%%%%%%%%%%%%%%%%%%%%%%%%
\section{Conclusion}
This work presented a novel adversarial architecture for the task of partial domain adaptation. The proposed model adopts a multi-class adversarial loss function to jointly align the marginal and class-conditional distributions across the shared classes between the source and target domains. Furthermore, it leverages two regularization functions to reduce the adverse effects of the outlier classes and the irrelevant samples in transferring information. Several experiments performed on the standard benchmark datasets for partial domain adaptation have demonstrated that our method can outperform the state-of-the-art methods on multiple adaptation tasks in terms of the classification performance.
%%%%%%%%%%%%%%%%%%%%%%%%%%%%%%%%%%%%%%%%%%
%%%%%%%%%%%%%%%%%%%%%%%%%%%%%%%%%%%%%%%%%%
\bibliographystyle{IEEEtran}
\bibliography{myegbib}

\end{document}

%% file: Notation.tex
\newcommand{\abf}{\boldsymbol{a}}
\newcommand{\Abf}{\boldsymbol{A}}

\newcommand{\Ccal}{\mathcal{C}}

\newcommand{\dbf}{\boldsymbol{d}}

\newcommand{\Dcal}{\mathcal{D}}

\newcommand{\ebf}{\boldsymbol{e}}

\newcommand{\Lcal}{\mathcal{L}}

\newcommand{\Mbf}{\boldsymbol{M}}

\newcommand{\Ocal}{\mathcal{O}}

\newcommand{\Rbb}{\mathbb{R}}

\newcommand{\xbf}{\boldsymbol{x}}

\newcommand{\Xcal}{\mathcal{X}}

\newcommand{\ybf}{\boldsymbol{y}}

\newcommand{\gammabf}{\boldsymbol{\gamma}}

\newcommand{\thetabf}{\boldsymbol{\theta}}

\newcommand{\sm}[2]{\scaleto{#1\mathstrut}{#2pt}}

\newcommand{\PaperName}{CCPDA}

\ifx\theorem\undefined
\newtheorem{theorem}{Theorem}
\newenvironment{theorem*}{\par\noindent{\bf Theorem\ }}{\hfill\\[2mm]}

\newenvironment{corollary*}{\par\noindent{\bf Corollary\ }}{\hfill\\[2mm]}

\fi

\makeatletter
\DeclareRobustCommand{\cev}[1]{%
  \mathpalette\do@cev{#1}%
}
\newcommand{\do@cev}[2]{%
  \fix@cev{#1}{+}%
  \reflectbox{$\m@th#1\vec{\reflectbox{$\fix@cev{#1}{-}\m@th#1#2\fix@cev{#1}{+}$}}$}%
  \fix@cev{#1}{-}%
}
\newcommand{\fix@cev}[2]{%
  \ifx#1\displaystyle
    \mkern#2 1mu
  \else
    \ifx#1\textstyle
      \mkern#2 3mu
    \else
      \ifx#1\scriptstyle
        \mkern#2 2mu
      \else
        \mkern#2 2mu
      \fi
    \fi
  \fi
}

\crefmultiformat{table}{(#2#1#3)}{\,--\,(#2#1#3)}{#2#1#3}{#2#1#3}
\crefrangeformat{equation}{(#3#1#4)\,--\,(#5#2#6)}
\crefname{figure}{Figure}{Figures}
\crefname{equation}{}{}
\crefname{table}{Table}{Tables}
\crefname{lemma}{Lemma}{Lemmas}
\crefname{theorem}{Theorem}{Theorems}
\crefname{section}{Section}{Sections}
\crefname{subsection}{Subsection}{Sections}
\crefname{definition}{Definition}{Definitions}
\definecolor{darkblue}{rgb}{0.1,0.147,0.341}
\newcommand{\smallMinus}{\scalebox{1}[0.7]{-}}

\newcommand{\msh}{\!\!\:}
\newcommand{\tav}[1]{\scaleto{#1\mathstrut}{5.5pt}}

%% file: Page1-Figure.tex
\definecolor{Fcolor}{rgb}{0, 0.5, 0.25}
\definecolor{khakestari}{rgb}{0.94471,0.94471,0.94471}
\definecolor{khakestari2}{rgb}{0.84471,0.84471,0.84471}
\definecolor{ghermez}{rgb}{0.8392,0.1529,0.1569}
\definecolor{naranji}{rgb}{1,0.4980,0.0549}
\definecolor{abi}{rgb}{0.1216,0.4667,0.7059}
\definecolor{abi2}{rgb}{0,0.7490,0.7490}
\definecolor{naranji2}{rgb}{0.8,0.31,0.0}
\definecolor{sabz}{rgb}{0.1333, 0.6941, 0.2980}
\definecolor{khakestari3}{rgb}{0.24471,0.24471,0.24471}
\definecolor{Fcolor_1}{rgb}{0,0.447,0.741}
\definecolor{Fcolor_2}{rgb}{0.85,0.325,0.098}
\definecolor{Fcolor_3}{rgb}{0.929,0.694,0.125}
\definecolor{Fcolor_4}{rgb}{0.494,0.184,0.556}
\definecolor{Fcolor_5}{rgb}{0.466,0.674,0.188}
\definecolor{Fcolor_6}{rgb}{0.301,0.745,0.933}
\definecolor{Fcolor_7}{rgb}{0.635,0.078,0.184}

\begin{figure}
\centering
\begin{picture}(400,150)
% \begin{picture}(400,150)
% \put(0,0){\includegraphics[width=0.48\textwidth]{Fig_page1.pdf}}
% \put(11,32){\footnotesize {Outlier Class}}

% \put(4,125){\small Source Domain}
% \put(185,113){\small Target Domain}

% \put(105,121){\footnotesize {Shared Space}}
\put(0,0){\includegraphics[bb=0 0 200 100]{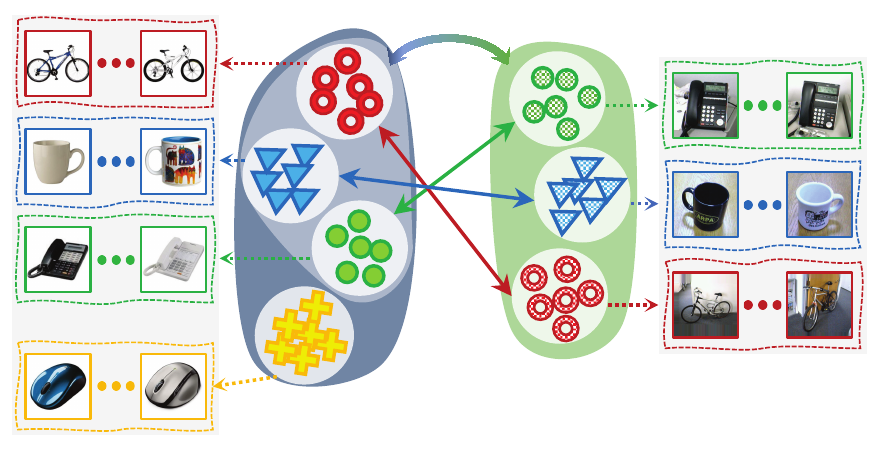}}
\put(11,33){\footnotesize {Outlier Class}}

\put(3,127){\small Source Domain}
\put(192,114){\small Target Domain}

\put(107,123){\footnotesize {Shared Space}}
%========================
\end{picture}
%%%%%%%%%%%%%%%%%%%%%%% \cite{georghiades2001few}
\caption{Illustration of partial domain adaptation task. The objective is to transfer knowledge between the shared classes in the source and target domains. To this end, it is desired to identify and reject the outlier source classes and align both marginal and class-conditional distributions across the shared label space. \textit{Best viewed in color.}}
\label{tb:Page1}
\end{figure}
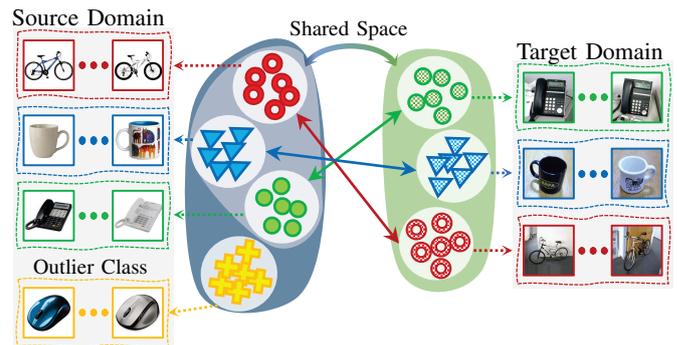

%% file: Model_fig.tex
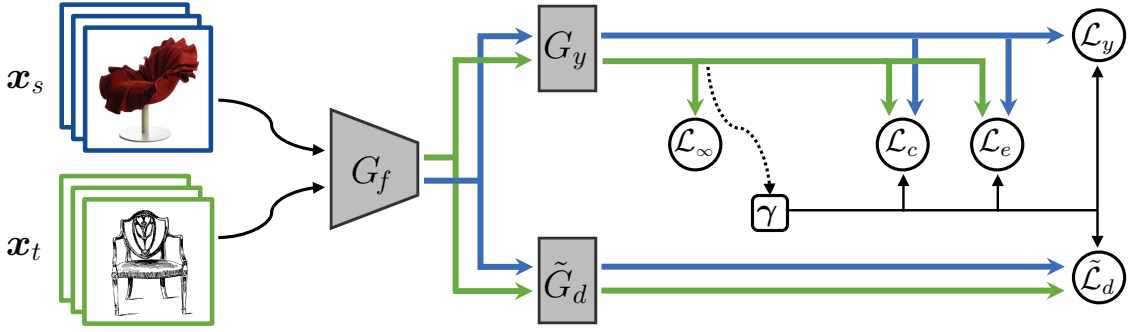
\begin{figure*}
\centering
\begin{picture}(400,150)
\put(0,0){\includegraphics[width=0.83\textwidth]{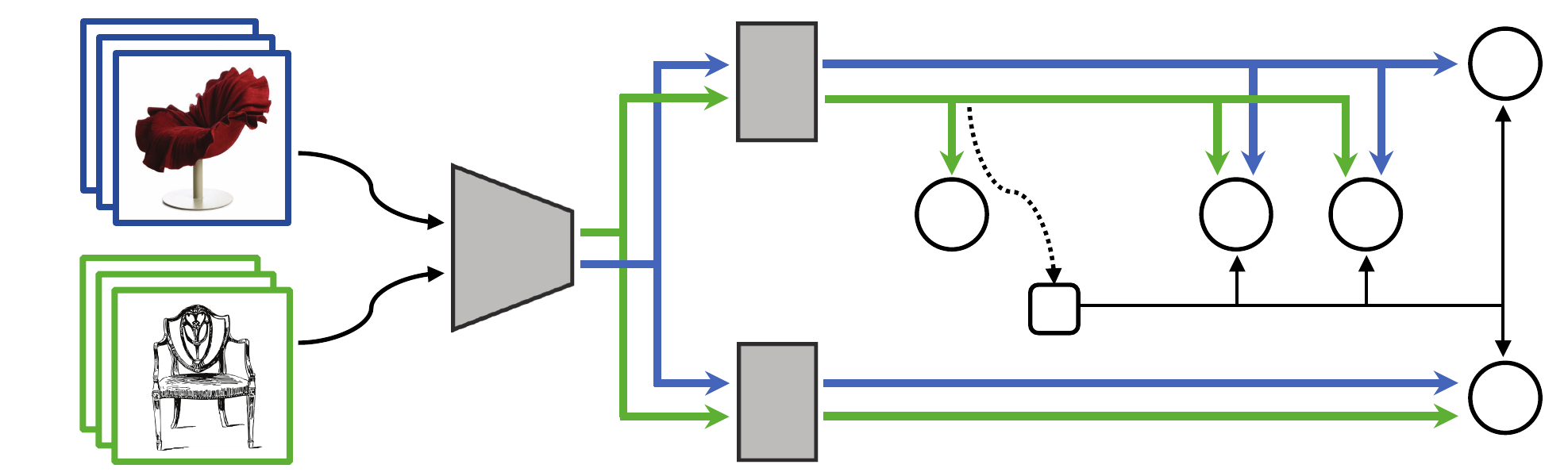}}
\put(131,57){\Large$G_{\!f}$}

\put(2,30){\Large$\xbf_{t}$}
\put(2,93){\Large$\xbf_{s}$}

\put(203,104){\Large$G_{y}$}
\put(203,14){\Large$\tilde{G}_{d}$}

\put(254,69){\large$\!\Lcal_{\sm{\infty}{6}}$}

\put(283,43){\large$\gammabf$}
\put(331,69){\large$\Lcal_{c}$}
\put(366,69){\large$\Lcal_{e}$}

\put(404,109){\large$\Lcal_{y}$}
\put(404,17){\large$\tilde{\Lcal}_{d}$}
%========================
\end{picture}
\caption{Overview of the proposed adversarial architecture for partial  transfer learning. The network consists of a feature extractor, a classifier, and a multi-class domain discriminator, denoted by $G_{\!f}$, $G_{\!y}$, and $\tilde{G}_{\!d}$, respectively. The blue arrows show the source flow and the green ones depict the target flow. Loss functions $\mathcal{L}_{y}$, $\tilde{\mathcal{L}}_{d}$, $\mathcal{L}_{c}$, $\mathcal{L}_{e}$, and $\mathcal{L}_{\scaleto{\infty}{2.5pt}}$ denote the classification loss, the discriminative loss, the centroid alignment loss, the entropy loss, and the selection loss, respectively. Parameter $\gammabf$ is computed based on the classifier output to target samples and then is used to weight the loss of different classes. \textit{Best viewed in color.}%
}
\label{fig:model}
\end{figure*}

%% file: Fig_office_home.tex
\begin{figure}
% \captionsetup[subfigure]{position=b}
\centering
%%%%%%%%%%%%%%%%%%%%%%%%%%%%%%%%%%%%%%%%%%%%%%%%%%%%%%%
	\begin{subfigure}[normal]{0.18\linewidth}
		\small{Art}
	\end{subfigure}
	\hspace{-0.0cm}
	\begin{subfigure}[normal]{0.66\linewidth}
		\includegraphics[scale=0.48,bb=0 0 125 70]{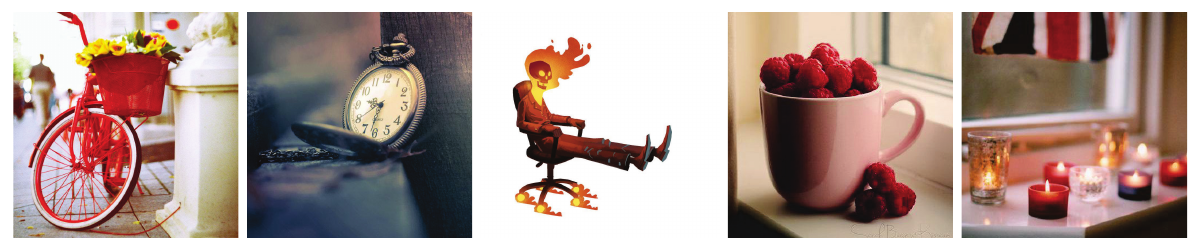}
	\end{subfigure}
	\\[0.0ex]
	\begin{subfigure}[normal]{0.18\linewidth}
		\small{Clipart}
	\end{subfigure}
	\hspace{-0.0cm}
	\begin{subfigure}[normal]{0.66\linewidth}
		\includegraphics[scale=0.48,bb=0 0 125 70]{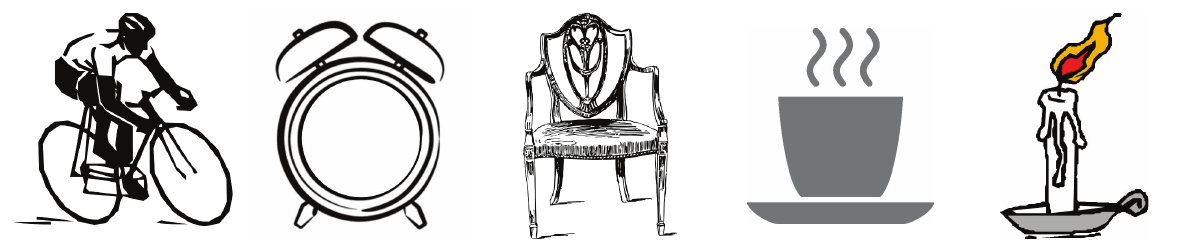}
	\end{subfigure}	
	\\[0.0ex]
    \begin{subfigure}[normal]{0.18\linewidth}
		\small{Real-World}
	\end{subfigure}
	\hspace{-0.0cm}
	\begin{subfigure}[normal]{0.66\linewidth}
		\includegraphics[scale=0.48,bb=0 0 125 70]{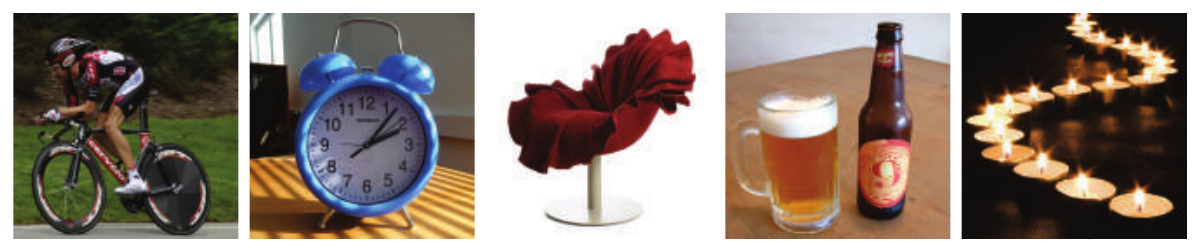}
	\end{subfigure}
	\\
	 \begin{subfigure}[normal]{0.18\linewidth}
		\small{Product}
	\end{subfigure}
	\hspace{-0.0cm}
	\begin{subfigure}[normal]{0.66\linewidth}
		\includegraphics[scale=0.48,bb=0 0 125 70]{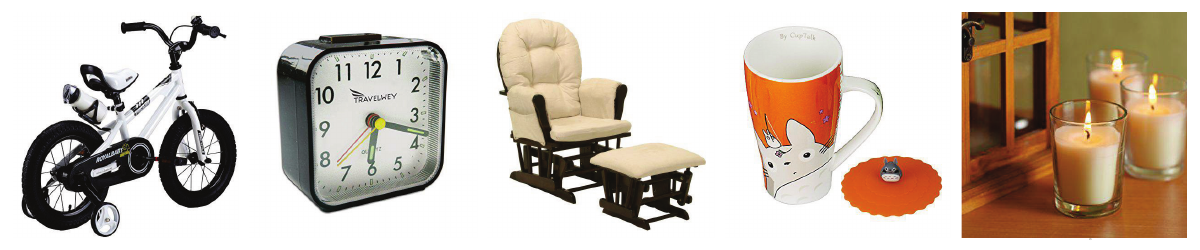}
	\end{subfigure}
\caption{Sample images from five classes in Office-Home dataset. Each column shows images from the same class but different domains.}
\label{fig:finalResults1}
\end{figure}

%% file: Table_office.tex
\begin{table*}[tbp]
    \addtolength{\tabcolsep}{3pt}
    \centering 
    \caption{Classification accuracy of partial domain adaptation tasks on Office-31.}
    \resizebox{0.66\textwidth}{!}{%
    \begin{tabular}{l|cccccc|c}
    \toprule[\heavyrulewidth]
         \multicolumn{1}{l}{Method} & \multicolumn{1}{c}{\textbf{A} $\rightarrow$ \textbf{W}} & \multicolumn{1}{c}{\textbf{D} $\rightarrow$ \textbf{W}} & \multicolumn{1}{c}{\textbf{W} $\rightarrow$ \textbf{D}} & \multicolumn{1}{c}{\textbf{A} $\rightarrow$ \textbf{D}} & \multicolumn{1}{c}{\textbf{D} $\rightarrow$ \textbf{A}} & \multicolumn{1}{c}{\textbf{W} $\rightarrow$ \textbf{A}} & \multicolumn{1}{c}{\textbf{Avg}} \\
        \midrule[\lightrulewidth]
        ResNet & 75.59 & 96.27 & 98.09 & 83.44 & 83.92 & 84.97 & 87.05\\[.5mm]
        DANN & 73.56 & 96.27 & 98.73 & 81.53 & 82.78 & 86.12 & 86.50  \\[.5mm]
        ADDA & 75.67 & 95.38 & 99.85 & 83.41 & 83.62 & 84.25 & 87.03 \\[.5mm]
        MADA & 90.00 & 97.40 & 99.60 & 87.80 & 70.30 & 66.40 & 85.20 \\[.5mm]
        RTN & 78.98 & 93.22 & 85.35 & 77.07 & 89.25 & 89.46 & 85.56 \\[.5mm]
        IWAN &  89.15 & 99.32 & 99.36 & 90.45 & 95.62 & 94.26 & 94.69\\[0.5mm] 
        SAN &  93.90 & 99.32 & 99.36 & 94.27 & 94.15 & 88.73 & 94.96\\[0.5mm] 
        PADA & 86.54 & 99.32 & \textbf{100.0} & 82.17 & 92.69 & 95.41 & 92.69 \\[0.5mm]
        ETN & 94.52 & \textbf{100.0} & \textbf{100.0} & 95.03 & \textbf{96.21} & 94.64 & 96.73 \\ [0.5mm] 
        \midrule[\lightrulewidth]
        {\PaperName} & \textbf{99.66} &  \textbf{100.0} & \textbf{100.0} & \textbf{97.45} & 95.72 & \textbf{95.71} & \textbf{98.09}
        \\
    \bottomrule
    \end{tabular}%
    }
    \label{tab:Office_31}
\end{table*}

%%%%%%%%%%%%%%%%%%%%%%%%%%%%%%
\begin{table*}
\centering 
\caption{Classification accuracy of partial domain adaptation tasks on Office-Home.}
\resizebox{\textwidth}{!}{%
\begin{tabular}{l|cccccccccccc|c}
    \toprule[\heavyrulewidth]
    \multicolumn{1}{l}{Method}
    & \multicolumn{1}{c}{\textbf{Ar}$\rightarrow$\textbf{Cl}} & \multicolumn{1}{c}{\textbf{Ar}$\rightarrow$\textbf{Pr}} & \multicolumn{1}{c}{\textbf{Ar}$\rightarrow$\textbf{Rw}} & \multicolumn{1}{c}{\textbf{Cl}$\rightarrow$\textbf{Ar}} & \multicolumn{1}{c}{\textbf{Cl}$\rightarrow$\textbf{Pr}} & \multicolumn{1}{c}{\textbf{Cl}$\rightarrow$\textbf{Rw}} & \multicolumn{1}{c}{\textbf{Pr}$\rightarrow$\textbf{Ar}} & \multicolumn{1}{c}{\textbf{Pr}$\rightarrow$\textbf{Cl}} & \multicolumn{1}{c}{\textbf{Pr}$\rightarrow$\textbf{Rw}} & \multicolumn{1}{c}{\textbf{Rw}$\rightarrow$\textbf{Ar}} & \multicolumn{1}{c}{\textbf{Rw}$\rightarrow$\textbf{Cl}} & \multicolumn{1}{c}{\textbf{Rw}$\rightarrow$\textbf{Pr}} & \multicolumn{1}{c}{\textbf{Avg}} \\
    \midrule[\lightrulewidth]

    %====================
    ResNet & 46.33 & 67.51 & 75.87 & 59.14 & 59.94 & 62.73 & 58.22 & 41.79 & 74.88 & 67.40 & 48.18 & 74.17 & 61.35 \\[.5mm]
    %=====================
    DANN & 43.76 & 67.90 & 77.47 & 63.73 & 58.99 & 67.59 & 56.84 & 37.07 & 76.37 & 69.15 & 44.30 & 77.48 & 61.72 \\[.5mm]
    %=====================
    ADDA & 45.23 & 68.79 & 79.21 & 64.56 & 60.01 & 68.29 & 57.56 & 38.89 & 77.45 & 70.28 & 45.23 & 78.32 & 62.82 \\[.5mm]
    %=====================
    RTN & 49.31 & 57.70 & 80.07 & 63.54 & 63.47 & 73.38 & 65.11 & 41.73 & 75.32 & 63.18 & 43.57 & 80.50 & 63.07 \\[.5mm]
    %=====================
    IWAN & 53.94 & 54.45 & 78.12 & 61.31 & 47.95 & 63.32 & 54.17 & 52.02 & 81.28 & 76.46 & 56.75 & 82.90 & 63.56 \\[.5mm]
    %=====================
    SAN & 44.42 & 68.68 & 74.60 & 67.49 & 64.99 & \textbf{77.80} & 59.78 & 44.72 & 80.07 & 72.18 & 50.21 & 78.66 & 65.30 \\
    %=====================
    PADA &51.95 & 67.00 & 78.74 & 52.16 & 53.78 & 59.03 & 52.61 & 43.22 & 78.79 & 73.73 & 56.60 & 77.09 & 62.06 \\[.5mm]
    % ====================
    ETN & \textbf{59.24} & 77.03 & 79.54 & 62.92 & 65.73 & 75.01 & 68.29 & \textbf{55.37} & \textbf{84.37} & 75.72 & \textbf{57.66} & 84.54 & 70.45 \\[.5mm]
    % ====================
     \midrule[\lightrulewidth]
    {\PaperName}  & 55.31 & \textbf{80.11} & \textbf{88.07} & \textbf{73.28} & \textbf{71.21} & 77.63 & \textbf{71.89} & 52.97 & 81.41 & \textbf{81.81} & 56.21 & \textbf{85.15} & \textbf{72.92} \\
    \bottomrule
\end{tabular}%
}
\label{tab:office_home}
\end{table*}

%% file: Table_caltech_office.tex
\begin{table}[ht]
    \addtolength{\tabcolsep}{3pt}
    \centering 
    \caption{Classification accuracy of partial domain adaptation tasks on Caltech-Office.}
    \resizebox{0.38\textwidth}{!}{%
    \begin{tabular}{l|@{\!}c@{\!}c@{\!}c@{\!}|c}
    \toprule[\heavyrulewidth]
         \multicolumn{1}{l}{Method} & \multicolumn{1}{c}{\textbf{C} $\rightarrow$ \textbf{W}} & \multicolumn{1}{c}{\textbf{C} $\rightarrow$ \textbf{A}} & \multicolumn{1}{c}{\textbf{C} $\rightarrow$ \textbf{D}} & \multicolumn{1}{c}{\textbf{Avg}} \\
        \midrule[\lightrulewidth]
        {RevGrad} & 54.57 & 72.86 & 57.96 & 61.80 \\
        {ADDA} & 73.66 & 78.35 & 74.80 & 75.60 \\
        {RTN} & 71.02 & 81.32 & 62.35 & 71.56 \\
        {SAN} & 88.33 & 83.82 & 85.35 & 85.83 \\
        % ======================
        \midrule[\lightrulewidth]
        {\PaperName} & \textbf{95.23} & \textbf{88.05} & \textbf{100.0} & \textbf{94.42} \\
    \bottomrule
    \end{tabular}%
    }
    \label{tab:caltech_Office}
\end{table}

%% file: tsne_fig.tex
%%%%%%%%%%%%%%%%%%%%%%%%%%%%%%%%%%%%%%%%%%%%%%%%
\definecolor{Fcolor}{rgb}{0, 0.5, 0.25}
\definecolor{khakestari}{rgb}{0.9725,0.4549,0.5451}
\definecolor{sabz2}{rgb}{0.7098,0.9019,0.1137}
\definecolor{ghermez}{rgb}{0.8392,0.1529,0.8569}
\definecolor{naranji}{rgb}{1,0.4980,0.0549}
\definecolor{abi}{rgb}{0.1216,0.4667,0.7059}
\definecolor{abi2}{rgb}{0,0.7490,0.7490}
\definecolor{naranji2}{rgb}{0.8,0.31,0.0}
\definecolor{sabz}{rgb}{0.1333, 0.6941, 0.2980}
%%%%%%%%%%%%%%%%%%%%%%%%%%%%%%%%%%%%%%%%%%%%%%%%
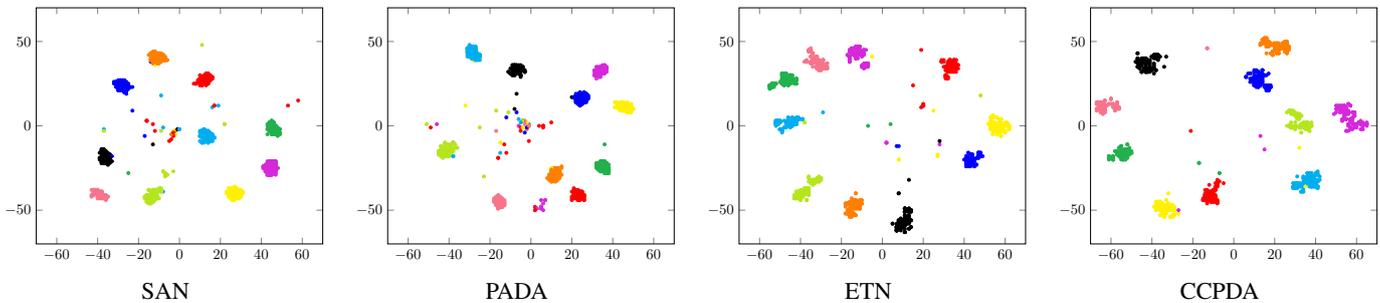
\begin{figure*}
% \captionsetup[subfigure]{margin={0.7cm,0cm}}
% \centering
\hspace{-0.2cm}
\begin{subfigure}[normal]{0.24\linewidth}
% \centering
\begin{tikzpicture}[scale=0.55]
\centering
\begin{axis}[
% xlabel = {\LARGE{First Dimension}},
% ylabel = {\LARGE{Second Dimension}},
% ylabel near ticks,
% xlabel near ticks,
xmax = 70,
xmin = -70,
ymax = 70,
ymin = -70,
scatter/classes={
    0={mark=*,draw=sabz,sabz,mark size=1.0pt},
    1={mark=*,draw=khakestari,khakestari,mark size=1.0pt},
    2={mark=*,draw=ghermez,ghermez,mark size=1.0pt},
    3={mark=*,draw=blue,blue,mark size=1.0pt},
    4={mark=*,draw=yellow,yellow,mark size=1.0pt},
    5={mark=*,draw=cyan,cyan,mark size=1.0pt},
    6={mark=*,draw=sabz2,sabz2,mark size=1.0pt},
    7={mark=*,draw=orange,orange,mark size=1.0pt},
    8={mark=*,draw=black,black,mark size=1.0pt},
    9={mark=*,draw=red,red,mark size=1.0pt}
  }
]
\addplot[scatter,only marks,scatter src=explicit symbolic]
    table[meta = label] {./data_san_1.tex};
\end{axis}
\end{tikzpicture}
\caption*{\small SAN}
\end{subfigure}
\hspace{0.05cm}
%%%%%%%%%%%%%%%%%%%%%%%%%%%%
\begin{subfigure}[normal]{0.24\linewidth}
% \centering
\begin{tikzpicture}[scale=0.55]
\centering
\begin{axis}[
% xlabel = {\LARGE{First Dimension}},
% ylabel = {\LARGE{Second Dimension}},
% ylabel near ticks,
% xlabel near ticks,
xmax = 70,
xmin = -70,
ymax = 70,
ymin = -70,
scatter/classes={
    0={mark=*,draw=sabz,sabz, mark size=1.0pt},
    1={mark=*,draw=khakestari,khakestari, mark size=1.0pt},
    2={mark=*,draw=ghermez,ghermez, mark size=1.0pt},
    3={mark=*,draw=blue,blue, mark size=1.0pt},
    4={mark=*,draw=yellow,yellow, mark size=1.0pt},
    5={mark=*,draw=cyan,cyan, mark size=1.0pt},
    6={mark=*,draw=sabz2,sabz2, mark size=1.0pt},
    7={mark=*,draw=orange,orange, mark size=1.0 pt},
    8={mark=*,draw=black,black, mark size=1.0 pt},
    9={mark=*,draw=red,red, mark size=1.0 pt}
  }
]
\addplot[scatter,only marks,scatter src=explicit symbolic]
    table[meta = label] {./data_pada_1.tex};
\end{axis}
\end{tikzpicture}
\caption*{\small PADA}
\end{subfigure}
\hspace{0.05cm}
%%%%%%%%%%%%%%%%%%%%%%%
\begin{subfigure}[normal]{0.24\linewidth}
% \centering
\begin{tikzpicture}[scale=0.55]
\centering
\begin{axis}[
% xlabel = {\LARGE{First Dimension}},
% ylabel = {\LARGE{Second Dimension}},
% ylabel near ticks,
% xlabel near ticks,
xmax = 70,
xmin = -70,
ymax = 70,
ymin = -70,
scatter/classes={
    0={mark=*,draw=sabz,sabz,mark size=1.0pt},
    1={mark=*,draw=khakestari,khakestari,mark size=1.0pt},
    2={mark=*,draw=ghermez,ghermez,mark size=1.0pt},
    3={mark=*,draw=blue,blue,mark size=1.0pt},
    4={mark=*,draw=yellow,yellow,mark size=1.0pt},
    5={mark=*,draw=cyan,cyan,mark size=1.0pt},
    6={mark=*,draw=sabz2,sabz2,mark size=1.0pt},
    7={mark=*,draw=orange,orange,mark size=1.0pt},
    8={mark=*,draw=black,black,mark size=1.0pt},
    9={mark=*,draw=red,red,mark size=1.0pt}
    % 9={mark=*,draw=red,red, mark size=1.0pt}
  }
]
\addplot[scatter,only marks,scatter src=explicit symbolic]
    table[meta = label] {./data_etn_1.tex};
\end{axis}
\end{tikzpicture}
\caption*{\small ETN}
\end{subfigure}
\hspace{0.05cm}
%%%%%%%%%%%%%%%%%%%%%
\begin{subfigure}[normal]{0.24\linewidth}
% \centering
\begin{tikzpicture}[scale=0.55]
\centering
\begin{axis}[
% xlabel = {\LARGE{First Dimension}},
% ylabel = {\LARGE{Second Dimension}},
% ylabel near ticks,
% xlabel near ticks,
xmax = 70,
xmin = -70,
ymax = 70,
ymin = -70,
scatter/classes={
    0={mark=*,draw=sabz,sabz, mark size=1.0pt},
    1={mark=*,draw=khakestari,khakestari,mark size=1.0pt},
    2={mark=*,draw=ghermez,ghermez,mark size=1.0pt},
    3={mark=*,draw=blue,blue,mark size=1.0pt},
    4={mark=*,draw=yellow,yellow,mark size=1.0pt},
    5={mark=*,draw=cyan,cyan,mark size=1.0pt},
    6={mark=*,draw=sabz2,sabz2,mark size=1.0pt},
    7={mark=*,draw=orange,orange,mark size=1.0pt},
    8={mark=*,draw=black,black,mark size=1.0pt},
    9={mark=*,draw=red,red,mark size=1.0pt}
  }
]
\addplot[scatter,only marks,scatter src=explicit symbolic]
    table[meta = label] {./data_us_1.tex};
\end{axis}
\end{tikzpicture}
\caption*{\small{\PaperName}}
\end{subfigure}
%%%%%%%%%%%%%%%%%%
\caption{The t-SNE visualization of SAN \cite{cao2018san}, PADA \cite{cao2018partial}, ETN \cite{cao2019learning}, and {\PaperName} on partial domain adaptation task \textbf{A}$\,\rightarrow\,$\textbf{W} with class information (samples are colored w.r.t. their classes). \textit{Best viewed in color.}}
\label{fig:tsne}
\end{figure*}
%%%%%%%%%%%%%%%%%%%%%%%%%%%%%%%%%%%%%%%%%%%%%%%%%%

%% file: Convergance_Table.tex
\definecolor{Fcolor}{rgb}{0, 0.5, 0.25}
\definecolor{khakestari}{rgb}{0.94471,0.94471,0.94471}
\definecolor{khakestari2}{rgb}{0.84471,0.84471,0.84471}
\definecolor{ghermez}{rgb}{0.8392,0.1529,0.1569}
\definecolor{naranji}{rgb}{1,0.4980,0.0549}
\definecolor{abi}{rgb}{0.1216,0.4667,0.7059}
\definecolor{abi2}{rgb}{0,0.7490,0.7490}
\definecolor{naranji2}{rgb}{0.8,0.31,0.0}
\definecolor{sabz}{rgb}{0.1333, 0.6941, 0.2980}
\definecolor{khakestari3}{rgb}{0.24471,0.24471,0.24471}
\definecolor{Fcolor_1}{rgb}{0,0.447,0.741}
\definecolor{Fcolor_2}{rgb}{0.85,0.325,0.098}
\definecolor{Fcolor_3}{rgb}{0.929,0.694,0.125}
\definecolor{Fcolor_4}{rgb}{0.494,0.184,0.556}
\definecolor{Fcolor_5}{rgb}{0.466,0.674,0.188}
\definecolor{Fcolor_6}{rgb}{0.301,0.745,0.933}
\definecolor{Fcolor_7}{rgb}{0.635,0.078,0.184}

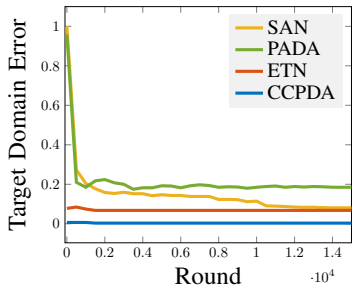
\begin{figure}
\centering
\captionsetup[subfigure]{margin={0.88cm,0cm}}
\hspace{-0.8cm}
%%%%%%%%%%%%%%%
\begin{subfigure}[normal]{0.5\textwidth}
\centering
\begin{tikzpicture}[scale = 0.55]
\begin{axis}[
xmin=-100,
xmax= 15000,
%%%%%%%%%% Ticks
ylabel near ticks,
xlabel near ticks,
xlabel = {\LARGE Round},
ylabel = {\LARGE Target Domain Error},
%%%%%% Legend
legend style={ font = \Large, draw=none, fill=khakestari},
% every tick label/.append style={font=\large},
legend pos = north east,
legend image post style={scale=1.1},
legend cell align={left},
]
%%%%% 3
\addplot[color=Fcolor_3, line width=0.75mm] table[x=x,y=y_s] {res.tex};
% %%%%% 4
\addplot[color=Fcolor_5, line width=0.75mm] table[x=x,y=y_p] {res.tex};
% %%%%% 5
\addplot[color=Fcolor_2, line width=0.75mm] table[x=x,y=y_e] {res.tex};
% %%%%% 6
\addplot[color=Fcolor_1, line width=0.75mm] table[x=x,y=y_u] {res.tex};
%%%%%%%%%%%%%%%%%%%%%%%%%%%%%%%%%%%%%
\legend{SAN,PADA,ETN,{\PaperName}}
\end{axis}
\end{tikzpicture}
\end{subfigure}
%%%%%%%%%%%%%%%%%%%%%%% \cite{georghiades2001few}
\caption{Empirical analysis of the target domain error through the training process. \textit{Best viewed in color.}}
\label{tb:convergance}
\end{figure}

%% file: Table_ablation.tex
%%%%%%%%%%%%%%%%%%%%%%%%%%%%%%
\begin{table}
\centering 
\caption{Classification accuracy of {\PaperName} and its variants for Partial Domain Adaptation tasks on Office-31 dataset.}
\resizebox{0.47\textwidth}{!}{%
\begin{tabular}{l|cccccc|c}
    \toprule[\heavyrulewidth]
    \multicolumn{1}{l}{Method}&
    \multicolumn{1}{c}{\textbf{A}$\rightarrow$\textbf{W}} & \multicolumn{1}{c}{\textbf{D}$\rightarrow$\textbf{W}} & \multicolumn{1}{c}{\textbf{W}$\rightarrow$\textbf{D}} & \multicolumn{1}{c}{\textbf{A}$\rightarrow$\textbf{D}} & \multicolumn{1}{c}{\textbf{D}$\rightarrow$\textbf{A}} & \multicolumn{1}{c}{\textbf{W}$\rightarrow$\textbf{A}} & \multicolumn{1}{c}{\textbf{Avg}} \\
    \midrule[\lightrulewidth]
    %=====================
    % $\text{\PaperName}_{\sm{\infty}{6},e}$  & 94.28 & 98.63 & -- & 94.90 & -- & -- & --
    % \\[.5mm]
    PADA & 86.54 & 99.32 & \textbf{100.0} & 82.17 & 92.69 & 95.41 & 92.69 \\[0.5mm]
    %=====================
    $\text{\PaperName}_{\sm{\infty}{6}}$  & 95.12 & 99.32 & \textbf{100.0} & 93.21 & \textbf{96.03} & 95.19 & 96.48 
    \\[.5mm]
    %====================
    $\text{\PaperName}_{e}$  & 97.45 & 96.64 & \textbf{100.0} & 96.47 & 94.92 & 93.86 & 96.56
    \\[.5mm]
    %=====================
    $\text{\PaperName}_{d,c}$ & 93.42 & 97.62 & \textbf{100.0} & 90.43 & 93.45 & 95.53 & 95.07
    \\[.5mm]
    % ====================
    \midrule[\lightrulewidth]
    $\text{\PaperName}$  & \textbf{99.66} & \textbf{100.0} & \textbf{100.0} & \textbf{97.45} & 95.72 & \textbf{95.71} & \textbf{98.09} \\
    \bottomrule
\end{tabular}%
}
\label{tab:ablation_study}
\end{table}